\pdfoutput=1

\documentclass[11pt]{article}

\usepackage[final]{acl}

\usepackage{times}
\usepackage{latexsym}

\usepackage{ulem}
\usepackage{colortbl}
\usepackage[most]{tcolorbox}
\newtcolorbox[auto counter, number freestyle={\noexpand\arabic{\tcbcounter}}]{definedbox}[2][]{%
    enhanced,
    colback=black!5!white,
    colframe=black!75!white,
    title=Example~\thetcbcounter: #2,
    #1
}
\usepackage{algorithm}
\usepackage[noend]{algorithmic}
\usepackage{listings}
\lstdefinelanguage{smt}{
    basicstyle=\small\ttfamily,
    morekeywords={Solver, add, from, import, def, return, Int},   
    stringstyle=\color{black},   
    breaklines=true,
    showstringspaces=false,
    tabsize=2
}

\usepackage[T1]{fontenc}

\usepackage[utf8]{inputenc}
\usepackage[T1]{fontenc}    
\usepackage{hyperref}       
\usepackage{url}            
\usepackage{booktabs}       
\usepackage{amsfonts}       
\usepackage{nicefrac}       
\usepackage{microtype}      
\usepackage{xcolor}         

\usepackage{bbding}
\usepackage{threeparttable}
\usepackage{xspace}
\usepackage{enumerate}
\usepackage{multirow}
\usepackage{multicol}
\usepackage{amsmath}

\usepackage{amssymb}
\usepackage{mathtools}
\usepackage{amsthm}
\usepackage{multirow}
\usepackage{colortbl}
\usepackage{pifont}
\usepackage{subfigure}
\usepackage{microtype}
\usepackage{tikz}

\usepackage{inconsolata}

\usepackage{graphicx}
\usepackage{todonotes}

\title{VCSearch: Bridging the Gap Between Well-Defined and Ill-Defined Problems in Mathematical Reasoning}

\author{
  Shi-Yu Tian$^{1,2 *}$, Zhi Zhou$^{1} \thanks{Equal contribution.}$, Kun-Yang Yu$^{1,2}$, Ming Yang$^{1,2}$, Lin-Han Jia$^{1}$, \\ 
  \textbf{Lan-Zhe Guo$^{1,3 \dagger}$, Yu-Feng Li$^{1,2}\thanks{Corresponding author.}$} \\
  $^1${National Key Laboratory for Novel Software Technology, Nanjing University}\\
  $^2${School of Artificial Intelligence, Nanjing University} \\
  $^3${School of Intelligence Science and Technology, Nanjing University} \\
  \texttt{\{tiansy,zhouz,guolz,liyf\}@lamda.nju.edu.cn}
}

\newcommand{\benchmark}{\text{PMC}\xspace}
\newcommand{\algo}{\textsc{VCSearch}\xspace}

\begin{document}

\maketitle
\begin{abstract}
Large language models (LLMs) have demonstrated impressive performance on reasoning tasks, including mathematical reasoning. However, the current evaluation mostly focuses on carefully constructed benchmarks and neglects the consideration of real-world reasoning problems that present missing or contradictory conditions, known as ill-defined problems. To further study this problem, we develop a large-scale benchmark called \emph{\textbf{P}roblems with \textbf{M}issing and \textbf{C}ontradictory conditions} (\benchmark) containing over 5,000 validated ill-defined mathematical problems. Our preliminary experiments through \benchmark reveal two challenges about existing methods: (1) traditional methods exhibit a trade-off between solving accuracy and rejection capabilities, and (2) formal methods struggle with modeling complex problems. To address these challenges, We develop \emph{\textbf{V}ariable-\textbf{C}onstraint \textbf{Search}} (\algo), a training-free framework that leverages formal language to detect ill-defined problems, where a variable-constraint pair search strategy is incorporated to improve the modeling capability of formal language. Extensive experiments demonstrate that \algo improves the accuracy of identifying unsolvable problems by at least 12\% across different LLMs, thus achieving stronger robust mathematical reasoning ability.
\end{abstract}



\section{Introduction}

Large language models (LLMs) have demonstrated strong performance on various reasoning tasks, including commonsense~\citep{zhao2023llm}, quantitative~\citep{lewkowycz2022quantitative}, and visual reasoning~\citep{gupta2023visual}. 
Mathematical problem solving~\citep{Cobbe2021gsm8k} serves as a fundamental benchmark for evaluating LLMs' reasoning capabilities~\citep{ahn2024large}. 
Recent advances in prompt-based methods~\citep{wei2022cot,ye2024satlm} and fine-tuning approaches~\citep{yu2023metamath,li2024neurosymbolic} have significantly improved their mathematical reasoning capabilities.
%

\begin{figure*}[t]
    \centering
    \includegraphics[width=0.9\linewidth]{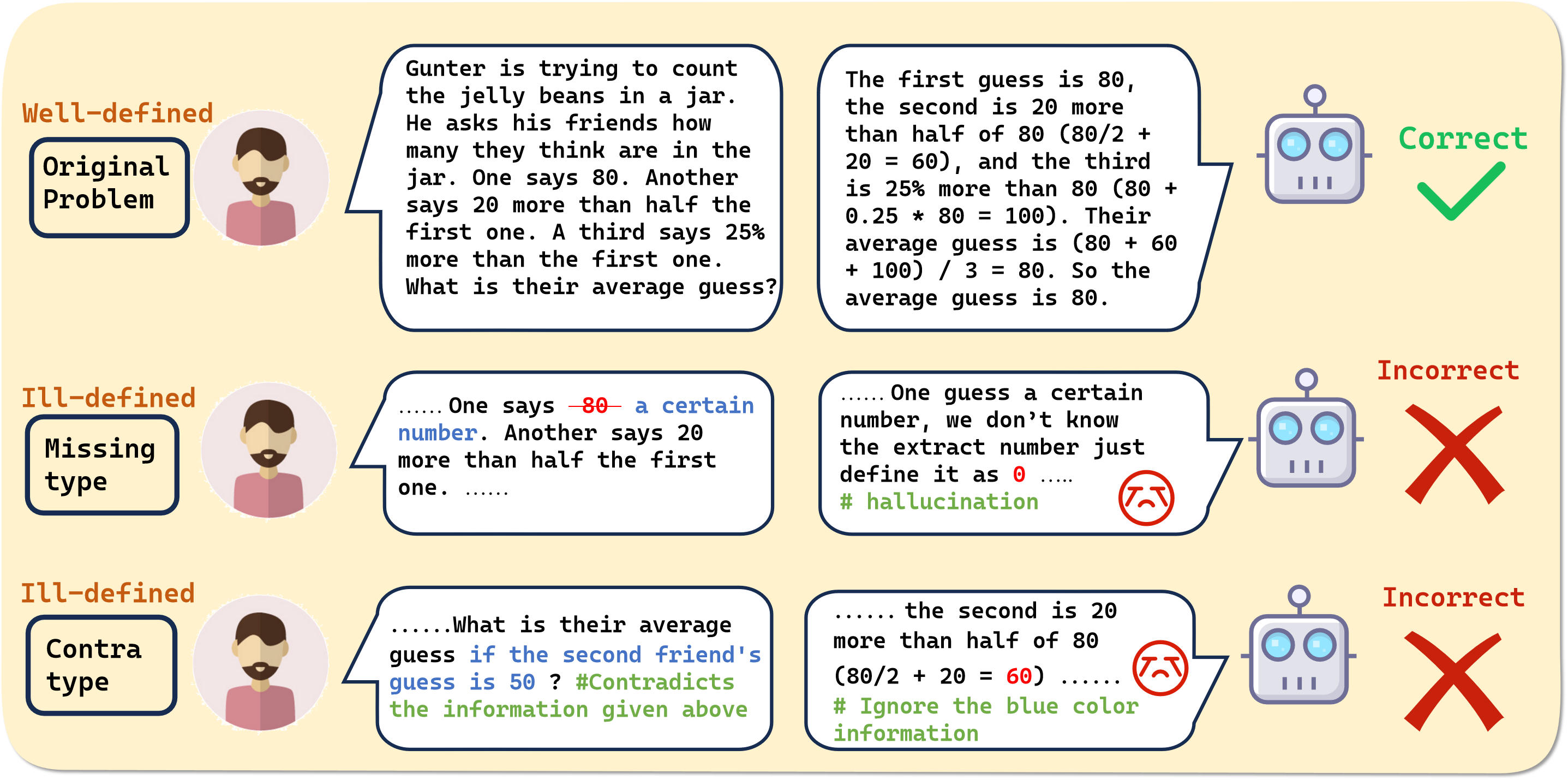}
    \caption{Well-defined problems and ill-defined
     problems in \benchmark with corresponding response. (Red strike-through indicates deleted sentences, blue indicates added sentences and green indicates explanation)}
     \vspace{-0.4cm}
    \label{fig:intro}
\end{figure*}

Although existing studies have improved the performance of LLMs on well-defined mathematical benchmarks~\citep{Cobbe2021gsm8k,patel2021svamp}, they often overlook a critical challenge in real-world applications: the ability to reject ill-defined problems~\citep{zhao2024mathtrap}. These problems, which contain missing or contradictory conditions~\citep{puchalska1987children}, are particularly common in educational scenarios. For instance, as shown in \autoref{fig:intro}, when students express mathematical problems unclearly, LLMs often generate plausible but incorrect solutions instead of identifying the problem as unsolvable. Such responses can reinforce misconceptions and hinder learning progress~\citep{ma2024unreasonmath}.

However, most existing benchmark about math reasoning robustness ~\cite{shi2023gsmic,zhou2024mathattack} focus on whether the model can still answer the question in the presence of interference, lacking a systematic evaluation of the model's ability to recognize and reject ill-defined problems. To better understand the limitations of existing methods and the development of novel mathematical reasoning methods, we build a large-scale evaluation dataset called \emph{\textbf{P}roblems with \textbf{M}issing and \textbf{C}ontradictory conditions} (\benchmark). This dataset contains over 5,000 validated ill-defined mathematical problems for comprehensive evaluation. 

Our preliminary experiments reveal two major challenges when handling ill-defined problems. First, traditional methods, e.g., prompt-based methods~\citep{yang2023alignmenthonesty} and fine-tuning approaches~\citep{zhao2024mathtrap}, demonstrate unsatisfactory performance due to an inherent trade-off between problem-solving accuracy and rejection capabilities. Second, although formal methods~\citep{ye2024satlm,pan2023logiclm,liu2024loT} offer unified problem-solving and rejection capabilities, they struggle to accurately model complex problems in formal language.


To address these challenges, we propose \algo (\emph{\textbf{V}ariable-\textbf{C}onstraint \textbf{Search}}), a training-free framework that systematically detects ill-defined problems through formal language to address the challenge of trade-offs. 
The key innovation of \algo lies in its variable-constraint dynamic search mechanism, which decomposes complex problems that are hard to model into dynamically extensible variable-constraint pairs, implementing an iterative optimization strategy where discovered variables guide constraint generation and existing constraints inform variable identification.
Experimental results demonstrate that \algo achieves an at least 12\% improvement in rejection accuracy for unsolvable problems compared to state-of-the-art methods, thus achieving stronger robust mathematical reasoning ability in realistic scenarios. 
Our main contributions can be summarized as follows:

\begin{enumerate}[1)]
    \item We introduce the practical challenge of evaluating robustness in mathematical reasoning and present \benchmark, a large-scale dataset comprising over 5,000 carefully validated ill-defined mathematical problems.
    \item We develop \algo, a training-free framework that leverages formal language to detect ill-defined problems, where a variable-constraint pair search strategy is incorporated to improve the modeling capability of formal language.
    \item Extensive experiments demonstrate that \algo consistently improves the accuracy of detecting unsolvable problems by over 12\% across multiple LLMs, establishing stronger and more reliable mathematical reasoning in practical settings.
\end{enumerate}

\section{\benchmark Benchmark and Analysis}
\label{sec:benchmark}

In this section, we first introduce our \benchmark benchmark, which consists of two types, i.e., Contra-type and Missing-type, by mutating problems from four common math datasets. 
Then, our analysis presents the challenges of rejecting ill-defined problems and the limitations of existing methods. 

\vspace{-0.2cm}
\subsection{Benchmark Construction}
\label{sec:dataset-construction}
We choose four common mathematical reasoning datasets, that is, GSM8k~\cite{Cobbe2021gsm8k}, SVAMP~\cite{patel2021svamp}, AddSub~\cite{hosseini2014AddSub}, and MultiArith~\cite{koncel2016multiarith}, as seed datasets to construct \benchmark. 
We define the problems in the seed dataset as \textbf{well-defined} problems, meaning that the given conditions in the problem statement are sufficient to derive a unique solution. In contrast, the problems we aim to construct are \textbf{ill-defined} problems, where the given conditions are insufficient—either due to missing necessary constraints or internal contradictions—making the problem unsolvable.

Our construction methodology employs a prompting-based strategy with Large Language Models (LLMs). Initially, the LLM is prompted to decompose a seed problem and ascertain all pertinent variables. Subsequently, the model is instructed to implement targeted modifications to the original problem conditions. To generate "missing-type" problems, a numerical value within a specific constraint is substituted with an indeterminate term, thereby rendering the problem definition incomplete. For "contra-type" problems, contradictory constraints pertaining to the variables are introduced, yielding problems that are inherently self-contradictory and thus pathological.
To verify the unsolvable (ill-defined) nature of the constructed problems, we utilize a panel of heterogeneous LLMs (e.g., Deepseek-V3~\cite{liu2024deepseek}, Doubao, and GLM~\cite{glm2024chatglm}) to assess whether the modified problem possesses a unique solution. A problem is classified as unsolvable if a consensus is reached among all participating LLMs that no solution exists. In instances where any model deems the problem solvable, human annotators are engaged to meticulously review the problem and confirm its unsolvable status.

Overall, \benchmark contains 8 different sub-datasets, including four Missing-type and four Contra-type datasets. An illustration of mutated problems of \benchmark is presented in Fig \ref{fig:intro}, and more detailed information about \benchmark (construction prompt, examples, etc.) can be found in Appendix~\ref{app:bench}.

\subsection{Evaluation Protocol}
\label{sec:evaluation-protocol}

To evaluate the robustness of methods in mathematical reasoning when faced with missing and contradictory conditions, we introduce two evaluation metrics: the Rejection Rate (R-Rate) and the Reaction Score (R-Score). 
R-Rate quantifies a method's ability to identify ill-defined problems. R-Score evaluates a method's overall performance in both handling ill-defined problems and solving well-defined problems.

For a well-defined dataset $\mathcal{D}_w$, let $\mathcal{D}_{i}$ be its ill-defined counterpart.  
For any problem $p$, let $g(p)$ denote its ground truth solution, where $g(p) = \mathrm{Reject}$ for ill-defined problems. 
Let $f(p)$ denote the solution generated by a method, where $f(p) = \mathrm{Reject}$ indicates the method rejects to solve $p$.
We define the R-Rate and R-Score as follows:

\paragraph{Rejection Rate.} Rejection Rate(R-Rate) is the percentage of ill-defined problems correctly rejected by method $f(\cdot)$: 
\begin{equation}
    \frac{ \sum_{p \in \mathcal{D}_i} \mathbb{I} \left [ f(p) = \mathrm{Reject} \right ]}{|\mathcal{D}_i|}
\end{equation}
\vspace{-0.2cm}

\paragraph{Reaction Score.} 
Reaction Score(R-Score) measures a method's overall performance by considering three scenarios:
(a) correctly rejecting ill-defined problems,
(b) correctly solving well-defined problems, and
(c) rejecting well-defined problems.
A method receives one point for scenarios (a) and (b), and 0.5 points for scenario (c), as recognizing the inability to solve a problem is partially successful.

\vspace{-0.2cm}
\begin{equation} 
\begin{aligned}
    (&\sum_{p \in \mathcal{D}_i}  \mathbb{I}[f(p) = \mathrm{Reject}]  
     + \sum_{p \in \mathcal{D}_{w}}   \mathbb{I}[ f(p) = g(p)]  \\ 
    & + 0.5  \sum_{p \in \mathcal{D}_{w}} \mathbb{I}[ f(p) = \mathrm{Reject} ])/ (|\mathcal{D}_i| + |\mathcal{D}_w|)
\end{aligned}
\end{equation}
\vspace{-0.2cm}


\subsection{Problem Analysis}
\begin{figure}[t]

\centering
\subfigure[ill-defined problems]{
\label{Fig.sub.1}
\includegraphics[width=3.5cm,height=3.5cm]{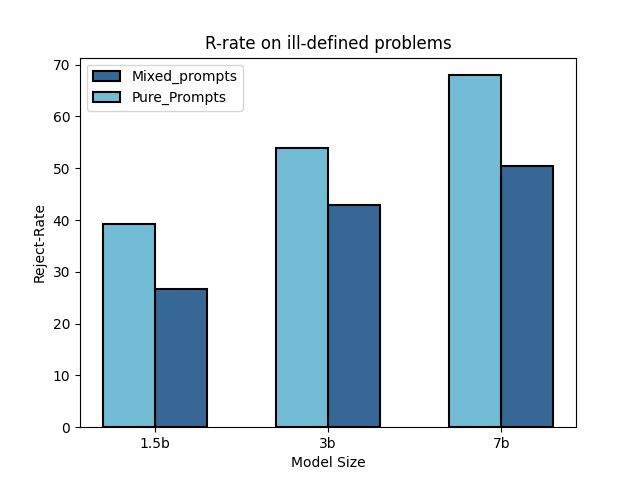}
}
\subfigure[well-defined problems]{
\label{Fig.sub.2}
\includegraphics[width=3.5cm,height=3.5cm]{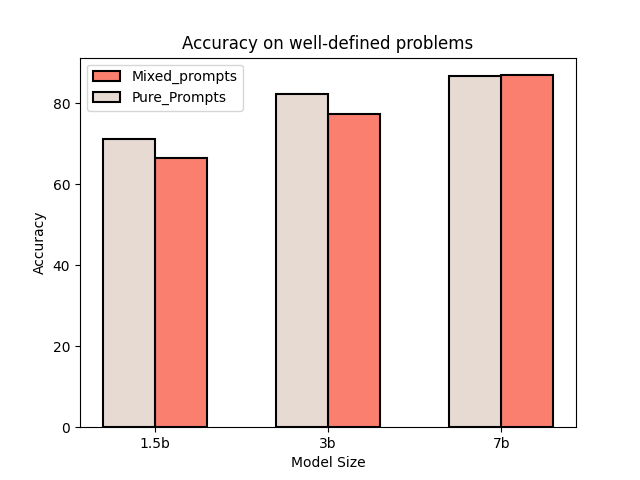}
}
\vspace{-0.3cm}
\caption{Trade-off faced by traditional methods when handling ill-defined and well-defined problems}
\label{fig:trade-off}
\vspace{-1.5em}
\end{figure}

We conduct a series of preliminary experiments on the \benchmark benchmark testing platform (with more detailed experimental modules to be elaborated in subsequent sections). The results are shown in Figure~\ref{fig:trade-off}. We use "pure prompt" to refer to directly prompting the model to solve well-defined or ill-defined problems (focusing on only one type), and "mixed prompt" to denote prompting the model to solve mathematical problems, where the model is instructed to reject if it deems the problem unsolvable. We observe that the base model exhibited certain problem-solving and rejection capabilities. However, there is a significant conflict between these two abilities: when the model is required to solve a problem while simultaneously employing a rejection mechanism, both its rejection and problem-solving capabilities are notably limited. This suggests a trade-off between the two abilities and this trade-off becomes more pronounced as the model size decreases.


\section{Methodology} 
\begin{figure*}[t]
    \centering
    \includegraphics[width=0.9\textwidth]{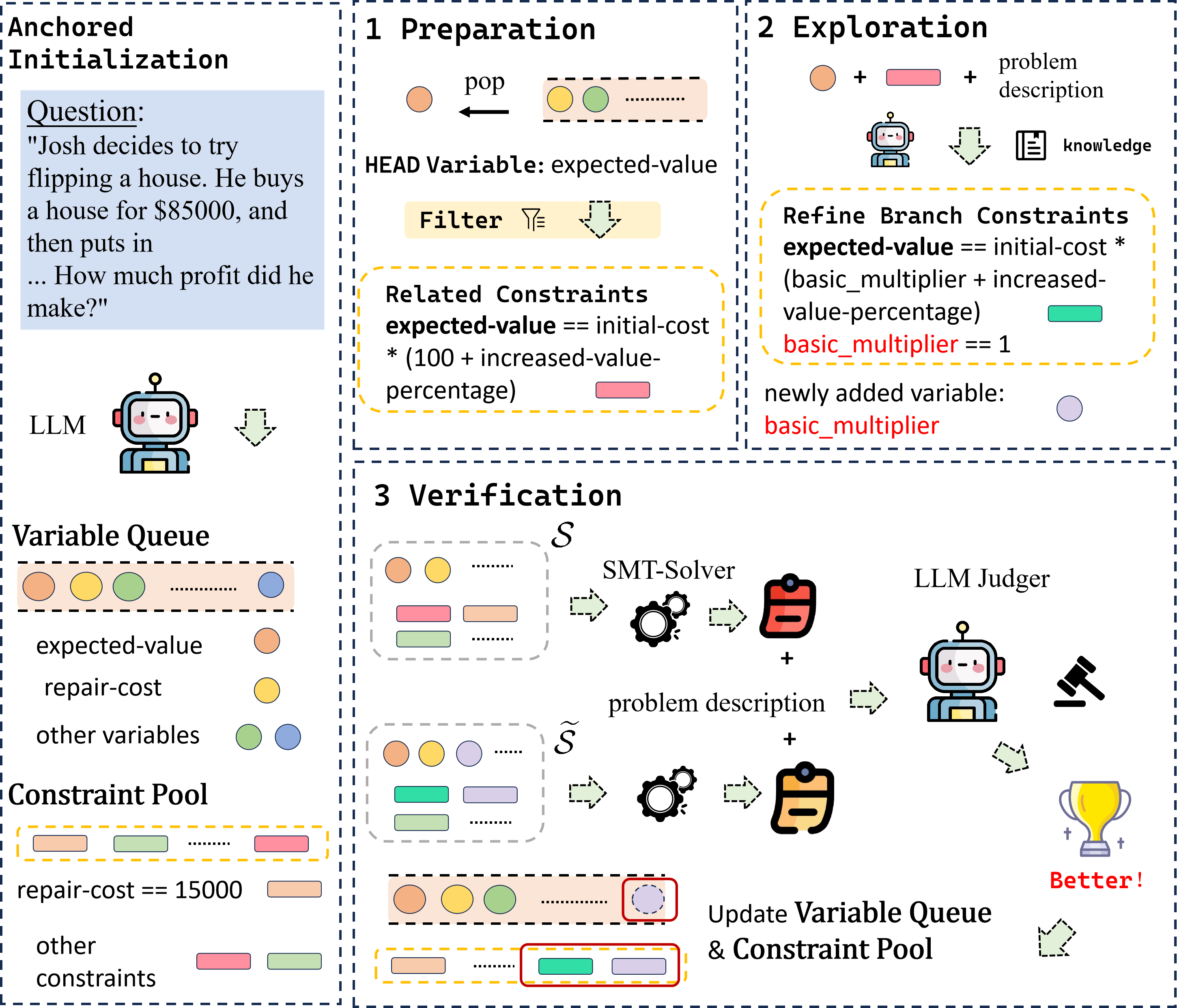}
    \caption{An overview of \algo. 
The left panel illustrates the outcome of a successful initialization phase,
culminating in an initialized draft formal modeling state, denoted as $S$.
Within this state representation, individual dots correspond to variables $v$,
while elongated rectangles signify constraints $c$.
Conversely, the right panel depicts the iterative process of \algo.
Each iteration commences with the extraction of a head variable,
followed by the sequential execution of three distinct steps:
(1)~Preparation, (2)~Exploration, and (3)~Verification.}
    \label{fig:frame}
    \vspace{-0.5em}
\end{figure*}

To mitigate the trade-off between solving accuracy and rejection capability, a natural idea is to incorporate formal solvers~\cite{ye2024satlm}. By leveraging their formal reasoning abilities, we can both detect ill-defined problems and augment existing mathematical reasoning methods with the capacity to recognize unsolvable cases.
However, modeling mathematical problems with formal language accurately is not trivial, directly using formalized examples as context prompts did not yield optimal results(in Table~\ref{Table:aood results}), raising the following challenge:
LLMs fail to model problems with formal language accurately in one pass. How can we improve the problem modeling ability? \\

To tackle this challenge, we first propose a \emph{Variable-Constraint Dynamic Search}(\algo) that systematically discovers new variables and constraints through an iterative searching process consisting of three steps: Preparation, Exploration, and Verification.
Then, to solve the cold start problem of search, we propose a \emph{Anchored Initialization} that leverages the reasoning capabilities of large models to reduce the initial search space. We use SMT-Lib~\cite{barrett2010smt} as the formal modeling language and Z3~\cite{Moura2008z3} as the formal solver in our approach and the overall framework is shown in Figure~\ref{fig:frame}.

\subsection{Variable-Constraint Dynamic Search}

LLMs have limitations in precisely modeling complex problems with formal language in a single pass due to the multiple variables and constraints involved which increase the modeling difficulty. 
We design a \emph{Variable-Constraint Dynamic Search} that decomposes complex problem modeling into a sequence of variable-constraint pair identification steps. This approach enables an iterative search that progressively improves the formal modeling. 

To achieve this, we implement the \emph{Variable-Constraint Dynamic Search} containing three systematic steps, i.e., Preparation, Exploration, and Verification. In each iteration, we perform the above three processes on the extracted variable.
For problem $p$, we denote the modeling state as $\mathcal{S} = (\mathcal{V}, \mathcal{C})$ where $\mathcal{V}$ is the set of variables and $\mathcal{C}$ is the set of constraints corresponding to $\mathcal{V}$. 


\paragraph{Preparation Step.}
This step selects a single variable and its associated constraints from $\mathcal{S}$ to reduce the complexity of the constraint analysis process, rather than considering all variables and constraints at once.
Given the variable set $\mathcal{V}$ and constraint set $\mathcal{C}$, we select one unexplored variable from the set $\mathcal{V}$ as the head variable $v_h$  and extract its related constraints $\mathcal{C}_h$ from $\mathcal{C}$:
\begin{equation}
    \mathcal{C}_h = \{c \mid v_h \in \text{vars}(c) \text{ and } c \in \mathcal{C}\}
\end{equation}
where $\text{vars}(\cdot)$ returns the set of variables in a given constraint, and $c$ represents a constraint from $\mathcal{C}$.

\paragraph{Exploration Step.}

This step explores new constraints and variables with the help of implicit knowledge from the LLM to improve the problem modeling. 
Specifically, we prompt the LLM to generate the polished constraints $\widetilde{\mathcal{C}}_h$, relating to variable $v_h$ for current problem $p$: 
\begin{equation}
    \widetilde{\mathcal{C}}_h = \text{LLM}_{E}(p, \, v_h , \, \mathcal{C}_h)
\end{equation}
where $\text{LLM}_{E}$ is denoted as the LLM prompted for exploration. The newly identified variables $\widetilde{\mathcal{V}}_h$ are
\begin{equation}
    \widetilde{\mathcal{V}}_h = \{v \mid v \in \text{vars}(\widetilde{\mathcal{C}}_h) \text{ and } v \notin \mathcal{V}\}. 
\end{equation}

\paragraph{Verification Step.} 

After exploring new constraints and variables, we can build a new problem modeling $\widetilde{\mathcal{S}}$ as follows. 
\begin{equation}
    \widetilde{\mathcal{S}} = \left(\mathcal{V} \cup \widetilde{\mathcal{V}}_h, \, (\mathcal{C} \setminus \mathcal{C}_h) \cup \widetilde{\mathcal{C}}_h\right)
\end{equation}
where the new variables are added at the tail of original variable set $\mathcal{V}$ and the polished constraints replaced the original related constraints in the constraint set $\mathcal{C}$.
Then, a SMT solver $\Phi$ is adopted to solve the problem modeling state $\widetilde{\mathcal{S}}$ and yield a solution $\widetilde{\mathcal{R}} = \Phi(\widetilde{\mathcal{S}})$. Inspired by LLMs as a judge~\citep{zheng2023llmjudging1,huang2024llmjudging2}, 
we compare the original problem modeling $\mathcal{S}$ with its solution $\mathcal{R}=\Phi(\mathcal{S})$ and the new problem modeling state $\widetilde{\mathcal{S}}$ with the solution $\widetilde{\mathcal{R}}$ as follows:
\begin{equation}
    \widetilde{\mathcal{S}}^{*} 
    = \text{LLM}_{J} \left ( p, (\mathcal{S}, \mathcal{R}), \, (\widetilde{\mathcal{S}}, \widetilde{\mathcal{R}}) \right )
\end{equation}
where $\text{LLM}_{J}$ is denoted as the LLM prompted for verification and $\widetilde{\mathcal{S}}^{*}$ is the selected state from new state $\widetilde{\mathcal{S}}$ and original state $\mathcal{S}$. 
Finally, we replace the current state $\mathcal{S}$ with the selected state $\widetilde{\mathcal{S}}^{*}$ for the subsequent process and add the newly detected variable to the variable queue $\mathcal{V}$. This repeated searching process is terminated until all variables in $\mathcal{V}$ are explored. 

This step not only ensures the adaptive nature of the search process but also effectively leverages the reasoning capabilities of LLMs to gradually improve problem modeling $\mathcal{S}$. In Appendix~\ref{app:algo}, we provide a specific example to illustrate the iterative process and provide detailed instructions and prompts for each step.



\subsection{Anchored Initialization}

However, the search process is particularly challenging at the outset due to the difficulty in initializing the search state, as the initial state contains limited information. The search space is vast, and without a reliable initialization, it is challenging to converge to a valid state. This can result in the model being overly conservative, leading to the rejection of many well-defined problems(Table~\ref{Table: ablation results}).

To address this challenge, we propose a \emph{Anchored Initialization} that leverages the reasoning capabilities of the LLM to generate a preliminary anchor state $\widehat{\mathcal{S}}$ as an anchored initialization state for \emph{Variable-Constraint Dynamic Search}.

Specifically, we first prompt the LLM to generate a draft modeling state $\hat{\mathcal{S}} = ( \widehat{\mathcal{V}}, \widehat{\mathcal{C}} )$ for problem $p$:
\begin{equation}
    ( \widehat{\mathcal{V}}, \widehat{\mathcal{C}} ) = \text{LLM}_{I}(p)
\end{equation}
where $\text{LLM}_{I}$ is denoted as the LLM prompted for initialization with four examples in the context. 
Then, we adopt a SMT solver $\Phi$ compute the solution $\widehat{\mathcal{R}} = \Phi(\widehat{\mathcal{S}})$ of the draft modeling state $\widehat{\mathcal{S}}$ for validation. 
If the solution $\widehat{\mathcal{R}}$ is valid, we regard the draft modeling state $\widehat{\mathcal{S}}$ as the initialization state $\mathcal{S}$ for \emph{Variable-Constraint Dynamic Search}. 
Otherwise, we only adopt the variable set $\widehat{\mathcal{V}}$ and empty constraint set as the initialization state $\mathcal{S}$ for subsequent searching. 
\begin{equation}
    \mathcal{S} = \begin{cases}
    (\widehat{\mathcal{V}}, \widehat{\mathcal{C}}) & \text{if } \Phi(\widehat{\mathcal{S}}) \neq \varnothing, \\
    (\widehat{\mathcal{V}}, \varnothing) & \text{if } \Phi(\widehat{\mathcal{S}}) = \varnothing.
    \end{cases}
\end{equation}
This module effectively incorporates the reasoning capabilities of the LLM to reduce the complexity of the search space at the beginning of the searching by providing a reliable initial anchor. 

\subsection{Integration with Existing Methods}

The \algo framework finally returns a problem modeling state $\mathcal{S}^{*} = (\mathcal{V}^{*}, \mathcal{C}^{*})$, and its solution can be computed by a SMT solver $\Phi$, i.e, $\mathcal{R}^{*} = \Phi(\mathcal{S}^{*})$.  
Therefore, we can integrate the \algo with any existing methods to enhance their ability to reject ill-defined problems. 
Specifically, we first verify the $\mathcal{R}^{*}$ set is valid by the \algo and the SMT solver. 
If $\mathcal{R}^{*}$ is valid, we regard the problem is well-defined and call existing methods to solve it.  Otherwise, we regard the problem is ill-defined and reject it.

In subsequent experiments, we report the performance of combining \algo with CoT~\cite{wei2022cot} and PAL~\cite{gao2023pal} to validate its effectiveness in practical applications.

\begin{table*}[t]
\caption{The rejection rates of various comparative methods on \benchmark}
\vspace{-1em}
\label{Table:aood results}
\resizebox{\textwidth}{!}{
\begin{tabular}{c|ccccc|ccccc}
\toprule
\bottomrule
\rowcolor[HTML]{FFCCC9}
\multicolumn{11}{l}{\bf Deepseek 6.7B} \\  
\hline
\multirow{2}{*}{Method} & \multicolumn{5}{c|}{\bf Contra-type}                 & \multicolumn{5}{c}{\bf Missing-type}                                     \\ \cline{2-11}
                        & Addsub & MultiArith & SVAMP & GSM8k & Avg  & Addsub & MultiArith & SVAMP & GSM8k & Avg  \\ \hline
Basic                   & ~~9.83  & 11.97      & 12.48 & ~~7.97 & 10.56 & ~~0.54  & ~~5.75   & ~~6.06 & ~~2.92 & ~~3.82                                    \\
CoT                     & 30.73  & 22.28      & 27.24 & 15.68 & 23.98 & 28.99  & 53.97  & 52.06 & 28.34 & 40.84                                        \\
PAL                     & ~~2.86  & ~~1.94      & ~~3.62 & ~~1.96 & ~~2.59 & ~~0.27  & ~~0.00  & ~~0.84 & ~~0.79 & ~~0.48                                              \\
Satlm                     & ~~5.73  & ~~2.78      & ~~4.83 & ~~6.79 & ~~5.03 & 68.83  & 63.28  & 64.36 & 46.04 & 60.63                                     \\
Ours                    & \textbf{ 54.09}  & \textbf{52.64}      & \textbf{54.89}
& \textbf{52.67} & \textbf{53.58} &  \textbf{89.70}     &         \textbf{88.49}             & \textbf{83.51}  & \textbf{63.68}    & \textbf{81.35}            \\
\hline
\rowcolor[HTML]{FFCCC9}
\multicolumn{11}{l}{\bf Qwen2.5 7B} \\  
\hline
\multirow{2}{*}{Method} & \multicolumn{5}{c|}{Contra-type}                 & \multicolumn{5}{c}{Missing-type}                                     \\ \cline{2-11}
                        & Addsub & MultiArith & SVAMP & GSM8k & Avg  & Addsub & MultiArith & SVAMP & GSM8k & Avg  \\ \hline
Basic                    & 27.86  & 22.00      & 25.23 & 28.36 & 25.86 & 79.94  & 75.97     & 80.24 & 64.57 & 75.18                                    \\
CoT                     & 36.88  & 31.75      & 44.69 & 38.16 & 37.87 & 71.27  & 80.54     & 82.18 & 55.09 & 72.27                                     \\
PAL                     & 47.54  & 42.06      & 46.57 & 41.96 & 44.53 & 82.11   &     89.34  &   91.51  &  82.22  & 79,97                                                 \\
Satlm                     & 12.29  & ~~9.47       & 16.24 & 23.79 &15.45 & 74.79  & 62.60                           & 66.06 & 44.10 &61.89                                        \\
Ours                    &   \textbf{48.36}  & \textbf{59.88}      & \textbf{56.44}
& \textbf{62.87} & \textbf{56.89} &  \textbf{97.01}   &          \textbf{95.93}              & \textbf{93.93} & \textbf{83.52}    &  \textbf{92.60}   \\
\hline
\rowcolor[HTML]{FFCCC9}
\multicolumn{11}{l}{\bf Qwen2.5 3B} \\ 
\hline
\multirow{2}{*}{Method} & \multicolumn{5}{c|}{Contra-type}                 & \multicolumn{5}{c}{Missing-type}                                     \\ \cline{2-11}
                        & Addsub & MultiArith & SVAMP & GSM8k & Avg  & Addsub & MultiArith & SVAMP & GSM8k & Avg  \\ \hline
Zero                        & 29.08                       & 23.39                & 34.22                & 28.75                & 28.86                      & 47.42                       & 54.99                      & 71.87 & 54.20 & 57.12                                          \\
CoT                         & 34.42                       & 36.21                & 42.01                & 30.06                & 35.67                      & 63.41                       & 73.09                      & 80.72 & 51.37 & 67.14                               \\
PAL                         & ~~3.28                        & ~~7.64                 & ~~5.90                 & 11.37                & ~~7.05                     &  17.07                           &     10.49                       &   26.67    &   17.18    &         17.85                                             \\
Satlm                      &  15.57                           &    ~~5.57                  &   16.24                   &      12.78               &        13.44                    & 54.74                       & 41.11                      & 43.39 & 26.73 & 41.49                                               \\
ours                        & \textbf{59.83}                      & \textbf{58.49}                & \textbf{60.00}             & \textbf{71.89}             & \textbf{62.53 }                  &    \textbf{93.49}                    &    \textbf{87.81}                 & \textbf{88.84}     &  \textbf{78.03}   &         \textbf{87.04}                                                 \\
\hline
\rowcolor[HTML]{FFCCC9}
\multicolumn{11}{l}{\bf Qwen2.5 1.5B} \\  
\hline
\multirow{2}{*}{Method} & \multicolumn{5}{c|}{Contra-type}                 & \multicolumn{5}{c}{Missing-type}                                     \\ \cline{2-11}
                        & Addsub & MultiArith & SVAMP & GSM8k & Avg  & Addsub & MultiArith & SVAMP & GSM8k & Avg  \\ \hline
Basic                    & 23.36  & 36.49    & 33.15& 26.92& 29.98 & 13.00  & 22.50    & 36.72& 20.72 & 23.23                                         \\
CoT                     &21.72  &32.59     &26.30 &25.35 & 26.49& 42.27  &51.60    & 59.63 & 45.17 & 49.67                                        \\
PAL                     & ~~4.91  & ~~7.52    & ~~6.04 & ~~9.80 & ~~7.06 & ~~4.06   &    ~~4.74  & ~~8.48 &   ~~6.83 &  ~~6.03                                              \\
Satlm                     & ~~6.55  & ~~3.06       & ~~7.91 & ~~6.27 & ~~5.94 & 27.91  & 19.12                           & 23.15 & 14.43& 21.15                                       \\
Ours                    &   \textbf{38.93 } & \textbf{32.59}      & \textbf{43.08}
& \textbf{40.91} & \textbf{38.87} & \textbf{73.44}  &            \textbf{63.41}              &  \textbf{64.48}  &\textbf{47.86}     &   \textbf{62.29}          \\
\bottomrule
\toprule
\end{tabular}
}
\end{table*}
\section{Experiments}
\label{sec:experiment}

In this section, we conduct experiments to answer the following three research questions.
\newline \textbf{RQ1.} Can \algo effectively identify and reject ill-defined problems?
\newline \textbf{RQ2.} Can \algo outperform formalized prompting method in modeling capabilities?
\newline \textbf{RQ3.} Can \algo help existing methods achieve robust mathematical reasoning in realistic scenarios?


\subsection{Experimental Setup}

\textbf{Datasets.} 
We conduct experiments on two types of datasets to validate our approach and address the three research questions: ill-defined problems and well-defined problems. For \textbf{ill-defined problems}, we primarily use our proposed \benchmark benchmark and Mathtrap~\cite{zhao2024mathtrap} dataset, which includes mathematical trap problems (Mathtrap results in Appendix). For \textbf{well-defined problems}, we utilize the original four subsets of \benchmark, which is AddSub~\cite{hosseini2014AddSub}, MultiArith~\cite{koncel2016multiarith}, SVAMP~\cite{patel2021svamp}, GSM8k~\cite{Cobbe2021gsm8k}, as well as Robustmath~\cite{zhou2024mathattack}, where symbols serve as interference signals, and GSM-IC~\cite{shi2023gsmic}, where irrelevant information serves as interference signals. 


\textbf{Compared methods.} 
We selected 4 well-behaved methods and compared them with our proposed \algo method. The methods are introduced as follows: 
(1)\textbf{Basic}, which is the zero-shot baseline method. 
(2)\textbf{CoT}, ~\cite{wei2022cot}, let model step-by-step reasoning before providing the final answer.
(3)\textbf{PAL}~\cite{gao2023pal}, modeling problem with python language.
(4)\textbf{Satlm}~\cite{ye2024satlm}, utilizes declarative prompting to model problems with satisfiability-aided language.

\begin{table*}[!t]
\centering
\caption{Comparison of the performance of Satlm and \algo on well-defined problems}
\vspace{-0.3cm}
\label{Table:aid results}
\begin{tabular}{l|cc|cc|cc|cc}
\toprule
\multirow{2}{*}{Dataset} & \multicolumn{2}{c|}{Deepseek 6.7B}  & \multicolumn{2}{c|}{Qwen 7B}         & \multicolumn{2}{c|}{Qwen 3B} & \multicolumn{2}{c}{Qwen 1.5B}\\ \cline{2-9} 
                         & Satlm & \multicolumn{1}{c|}{Ours} & Satlm & \multicolumn{1}{c|}{Ours} & Satlm & \multicolumn{1}{c|}{Ours}  & Satlm       & Ours       \\ \hline
Addsub                   &42.89&59.24            &72.15 &     85.31                      & 53.41          &  75.94   &    28.86   &  61.26     \\
MultiArith               & 73.50&72.50             &71.50     &      81.34                     &  39.50         &  59.67   &20.00&45.67      \\
SVAMP                    &50.21&  54.41           &70.80     &     82.10                      &  42.60        &  60.70      &18.70&40.80   \\
GSM8k                    &34.10&  41.31           &50.11     &     67.62                      &  29.34        &   41.31     & 10.32&21.37 \\
Robustmath               &44.33& 53.67            &55.33     &      75.67                    &  38.05         &   51.00      & 7.40 & 30.67  \\
GSM-IC                   &18.80&24.20              &49.20     &      74.52                     &  22.60         &  39.24     &  5.32 &12.00  \\ \hline
Avg                      & 43.97&  50.87            & 61.51    &     77.76                      &  37.58         &    54.64  & 15.10&  35.30    \\
\bottomrule
\end{tabular}
\vspace{-0.3cm}
\end{table*}

\textbf{Implementation Details.} 
Our main experiments are conducted on the Qwen2.5-Coder 7B/3B/1.5B~\cite{hui2024qwen25coder} and Deepseek-coder-6.7B~\cite{guo2024deepseekcoder}.
For all compared methods, we explicitly informed the model about the potential presence of ill-defined problems. Detailed settings and prompts can be found in the Appendix~\ref{app:expe}.

\subsection{Empirical Results}

\textbf{RQ1. Can \algo effectively identify and reject ill-defined problems?}

Our systematic evaluation on \benchmark (Table~\ref{Table:aood results}) shows that Contra-type tasks are substantially more challenging than Missing-type tasks, with all methods exhibiting lower performance. \algo achieved notable success on all ill-defined tasks, enabling the compared models to reach state-of-the-art performance and improving the rejection rate for identifying ill-defined problems by at least 12\% across different LLMs.
Further analysis indicates that the DeepSeek model struggled primarily because it tended to preset initial values (e.g., 0) for missing data, which hindered recognizability. By contrast, the Qwen series handled ill-defined problems more effectively, though its performance on long-context prompting was highly dependent on model scale. Distinctively, \algo exhibited strong robustness, maintaining consistent performance across models of varying sizes.


\noindent\textbf{RQ2. Can \algo outperform formalized prompting methods in modeling capabilities?}

In this section, we systematically compare \algo with traditional few-shot prompt methods that directly utilize the SMT-Lib language as in-context (Satlm). Since the ability to solve well-defined problems is a critical criterion for evaluating the modeling capabilities of algorithms, we focus on their performance in such tasks. The experimental results, presented in Table~\ref{Table:aid results}, demonstrate that \algo significantly outperforms conventional few-shot approaches. This underscores the effectiveness of the decomposition and search strategies introduced in our work, particularly for smaller base models, where these strategies lead to a substantial improvement in modeling capabilities. On average, accuracy improves by 14.95\%, with the most notable improvement observed in the Qwen 1.5B model, where accuracy increases from 15.10\% to 35.30\%. 
These findings show that \algo has effectively enhanced the model's ability to model problems.

\noindent\textbf{RQ3. Can \algo help existing methods achieve robust mathematical reasoning in realistic scenarios?}

In real-world scenarios, mathematical problems rarely fall into strictly well-defined or ill-defined categories. Instead, there is often a need to both solve well-defined problems and identify ill-defined ones. To the best of our knowledge, we are the first to explore this hybrid setting in the context of math word problems (MWP). For our experiments, we employed a balanced sampling strategy (e.g. $\mathcal{D}_w :\mathcal{D}_i =1:1 $) to fairly assess the ability to identify ill-posed problems and solve well-defined problems simultaneously.  This evaluation strategy is analogous to how imbalanced classification studies often report balanced metrics to properly assess model performance across all classes~\cite{thabtah2020imbalancesurvey}. After three repeated experiments, we report the mean ± standard deviation in Table \ref{Table: real results}.


The results show that \algo + CoT and \algo + PAL significantly outperform traditional CoT and PAL methods in rejecting unreasonable problems. The rejection rate of ill-defined problems improved by 42.96\% and 42.03\% respectively, while the real-world evaluation metrics R-score gained 16.78 and 19.39 points, confirming the application value of the hybrid architecture in complex real-world scenarios. We also provide additional discussions in the appendix, including a variation of the R-score metric and experimental results under different dataset proportions.

\begin{table}[t]
\centering
\caption{Reaction scores of \algo+ and comparison methods in a realistic environment with both ill-defined and well-defined problems}
\vspace{-0.5em}
\label{Table: real results}
\resizebox{\linewidth}{!}{
\begin{tabular}{c|l|cc}
\toprule
\multicolumn{1}{l|}{Model}   & Methods                                                              & R-Rate                                     & R-Score                                      \\ \hline
                             & CoT                                                                  &    51.33$\pm$2.29                                           &    65.93$\pm$0.73                                           \\
                             & \cellcolor[HTML]{DAE8FC}{\color[HTML]{333333}} +Ours & \cellcolor[HTML]{DAE8FC}{\color[HTML]{333333}} 76.13$\pm$1.56  & \cellcolor[HTML]{DAE8FC}{\color[HTML]{333333}} 73.98$\pm$0.28 \\
                             & PAL                                                                  &     14.46$\pm$0.41                                       &   48.56$\pm$0.22                                              \\
\multirow{-4}{*}{Qwen2.5 3B} & \cellcolor[HTML]{DAE8FC}{\color[HTML]{333333}} +Ours & \cellcolor[HTML]{DAE8FC}{\color[HTML]{333333}}75.59$\pm$1.39  & \cellcolor[HTML]{DAE8FC}{\color[HTML]{333333}}74.08$\pm$1.17  \\ \hline
                             & CoT                                                                  &   39.93$\pm$1.96                                              &                  53.91$\pm$1.16                               \\
                             & \cellcolor[HTML]{DAE8FC}{\color[HTML]{333333}} +Ours & \cellcolor[HTML]{DAE8FC}{\color[HTML]{333333}}65.06$\pm$1.48  & \cellcolor[HTML]{DAE8FC}{\color[HTML]{333333}}63.26$\pm$0.84  \\
                             & PAL                                                                  &~~7.73$\pm$2.04                                              &32.85$\pm$1.00                                            \\
\multirow{-4}{*}{Qwen2.5 1.5B} & \cellcolor[HTML]{DAE8FC}+Ours                        & \cellcolor[HTML]{DAE8FC} 66.66$\pm$0.24                       & \cellcolor[HTML]{DAE8FC}      62.28$\pm$0.65              \\  
\bottomrule
\end{tabular}
}
\end{table}
\vspace{-0.3cm}

\subsection{More discussion.}
\textbf{Ablations.}
In this part, we evaluate the impact of the two core components of \algo on overall performance (Table~\ref{Table: ablation results}). Removing the iterative search framework (i.e., replacing it with one-time refinement) yields only marginal improvement over the baseline SMT solver under few-shot learning. Excluding anchored initialization leads to severe search space divergence, causing the model to become overly conservative and reject most solutions, which substantially degrades its ability to solve well-defined problems. These results highlight the necessity of both components in this framework.


\begin{table}[]
\centering
\caption{Ablation study on Qwen 7B model.}
\label{Table: ablation results}
\vspace{-0.3cm}
\begin{small}
\begin{tabular}{cc|ccc}
\bottomrule
\toprule
Search & Initialization & R-Rate & Accuracy \\ 
\midrule
           & \checkmark &  43.59      &  61.28   \\
\checkmark &            &   89.97     &  22.81   \\
\checkmark & \checkmark &   74.75     &  77.76   \\
\bottomrule
\toprule
\end{tabular}
\end{small}
\vspace{-1.5em}
\end{table}

\begin{figure}[t]

\centering
\subfigure[ill-defined problems]{
\label{Fig.sub.1}
\includegraphics[width=3.5cm,height=3.5cm]{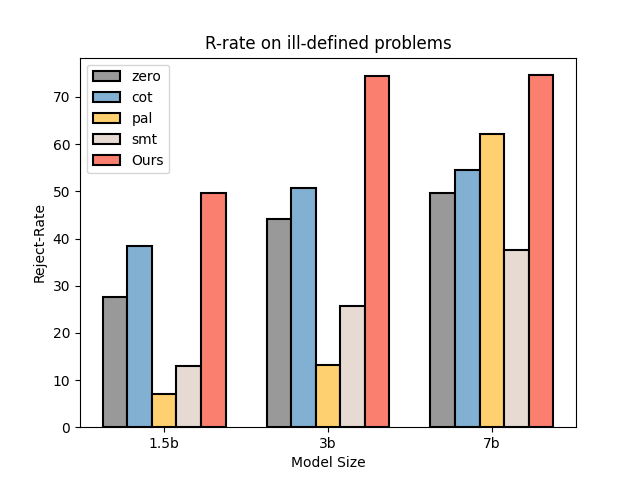}
}
\subfigure[well-defined problems]{
\label{Fig.sub.2}
\includegraphics[width=3.5cm,height=3.5cm]{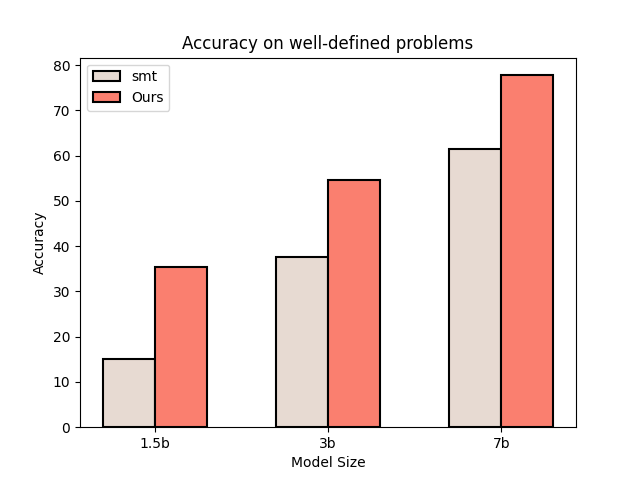}
}
\vspace{-1em}
\caption{Performance of \algo varying from different model size}
\label{fig:modelsize}
\vspace{-1.5em}
\end{figure}

\textbf{Performance of \algo on Models of Different Sizes.}
Visual analysis of Qwen model results (Figure~\ref{fig:modelsize}) reveals a strong correlation between model scale and performance: both ill-defined problem identification ability and well-defined problem solving ability decline with smaller models. However, our method mitigates this degradation and even shows advantages across scales. Specifically, \algo on Qwen-3B surpasses other methods on Qwen-7B in problem rejection and rivals SMT prompting on models an order of magnitude larger in solving well-defined problems, demonstrating its effectiveness and practical value in resource-limited scenarios.

\textbf{Miscellaneous.} 
In appendix~\ref{app:expe}, we include further discussions of experimental details, including the potential conservativeness of the r-score metric, the time efficiency of the proposed algorithm, as well as evaluations on additional benchmarks and more advanced large models, among others.

\section{Related work}
\label{sec:relatedwork}


\textbf{Enhancing Mathematical Reasoning in LLMs} Mathematical reasoning is a crucial aspect in evaluating model reasoning skills~\cite{xiong2025hsSTAR}, and there are currently two predominant lines for enhancing these skills. One line involves leveraging the existing few-shot prompt tool, such as CoT~\cite{wei2022cot}, PAL~\cite{gao2023pal}. The other is centered around fine-tuning strategy, like Metamath~\cite{yu2023metamath}, WizardMath~\cite{luo2023wizardmath} and Mugglemath~\cite{li2023mugglemath}.
Recent work has focused on how to achieve results that match or even exceed those of large models on smaller models~\cite{guan2025rstarmath} and smaller training datasets~\cite{zhi2024neurodata} by introducing techniques such as reinforcement learning and MCTS~\cite{tolpin2012mcts}.

\textbf{Robust Mathematical Reasoning} 
Model robustness is essential for secure deployment~\cite{sima2025viscra}, particularly in critical downstream applications such as finance~\cite{cao2022aifinace} or healthcare~\cite{tian2025ddi}. Traditional approaches, however, have mainly emphasized performance under noisy~\cite{liu2022labelnoise}, partially supervised~\cite{tian2024crosel}, or open-world settings~\cite{zhou2024decoop,zhou2025ftta}.
In recent years, there has been a significant surge in attention to the robustness of LLMs~\cite{morris2020advertextattack,wang2021adversarial}. In the context of robust mathematical reasoning, most existing work focuses on defining and constructing challenging "trap" datasets. For instance, Wang et.al ~\cite{wang2024mathood} treats mathematical problems from different datasets as an out-of-distribution (OOD) generalization problem.
Robustmath~\cite{zhou2024mathattack} introduces irrelevant punctuation marks as distractors, while GSMIC~\cite{shi2023gsmic} employs a sentence of unrelated contextual text to serve as a distractor, both aiming to investigate model performance variations. Recently, GSM-DC~\cite{yang2025gsmdc} analyzes how LLM reasoning is distracted by irrelevant context through controllable data generation.
The work most similar to ours is MathTrap~\cite{zhao2024mathtrap}, which focuses on a relatively small set of fewer than 300 ill-defined problems. In contrast, our PMC dataset is far more comprehensive, containing over 5,000 ill-defined problems.

\textbf{Neuro-Symbolic Methods with LLM reasoning.} 
Neuro-symbolic~\cite{colelough2025neuro,yang2025neurosurvey} methods have recently emerged as an effective approach to enhancing model reasoning capabilities and have been widely applied to reasoning augmentation and data generation across various downstream domains, including math~\cite{mirzadeh2024gsmsymbolic}, law~\cite{zhou2025lawgpt} and tabular data~\cite{tian2025autot2t}. 
The primary challenge of these methods lies in ensuring that the LLM correctly translates the reasoning problem from natural language (NL) to the formal language understood by the solver~\cite{raza2025logicchanllenge}. For instance, Logic-LM~\cite{pan2023logiclm} utilizes LLMs to convert natural language into symbolic formulas. SatLM~\cite{ye2024satlm} enables LLMs to generate task specifications that assist in translating natural language into logical predicates. LOT~\cite{liu2024loT}, similar to CoT, generates progressive logical paths. However, many of these methods struggle to extend successfully to smaller models, due to their limited contextual learning capabilities and lack of formal reasoning knowledge.

\section{Conclusion}

This paper addresses mathematical reasoning with missing and contradictory conditions by introducing \benchmark, a large-scale benchmark for evaluating LLM robustness. 
Our observations reveal a trade-off dilemma between reasoning for well-defined problems and recognizing ill-defined problems. To solve this trade-off, we propose \algo, a training-free framework that uses formal language to detect ill-defined problems, enhanced by a variable-constraint pair search strategy to improve formal modeling. Extensive experiments show \algo achieves superior robust reasoning across diverse model architectures and sizes.

\section*{Limitations}
Our work has two main limitations:

\noindent\textbf{Time Consumption.} Due to the use of variable-wise refinement and a search-based architecture during the reasoning process, our method inevitably incurs higher computational overhead compared to baseline approaches. While this additional cost is the price for improved robustness and broader applicability, it may limit scalability when applied to very large datasets or real-time scenarios. Future work may investigate techniques for reducing this overhead, such as pruning strategies or parallelization.  

\noindent\textbf{Limitations of Formal Tools.} Our method's ability to identify ill-defined problems heavily relies on formal tools, such as SMT solvers. By design, the system will directly reject tasks that cannot be adequately modeled with logical constraints. Although this ensures rigor in handling pathological cases, it may also lead to overly conservative behavior, including the incorrect rejection of certain well-defined problems. Extending the framework with more flexible or hybrid reasoning mechanisms could help alleviate this limitation.

\section*{Acknowledgements}
This research was supported by the Jiangsu Science Foundation (BG2024036, BK20243012), National Natural Science Foundation of China (624B2068,62576162,62576174), and the Fundamental Research Funds for the Central Universities (022114380023).


\bibliography{custom}

\newpage
\appendix
\onecolumn

\section{Appendix}
The appendix is organized as follows: Section A.1 provides additional details of our proposed dataset \benchmark; Section A.2 describes the operation process of our algorithm \algo; and Section A.3 presents further experiments and discussions.

\subsection{Details of benchmark \benchmark}
\label{app:bench}
We give more details of our constructed benchmark \benchmark here. 
\subsubsection{Composition and
examples of \benchmark}
We show the number of specific subsets of PMC in Table~\ref{table:number of pmc}, and show more representative problems to help understand our dataset.
\begin{table}[!htpb]
\centering
\caption{The specific number of rewritten datasets}
\label{table:number of pmc}
\begin{tabular}{cccccc}
\hline
Type                         & AddSub & MultiArith & SVAMP & GSM8k & Sum  \\ \hline
\multicolumn{1}{c}{M-type} & 369    & 591        & 825   & 1129  & 2914 \\ \hline
C-type                       & 244    & 359        & 745   & 765   & 2113 \\ \hline
\end{tabular}
\end{table}

\begin{figure}[thb]
    \label{fig:pmc example1}
    \begin{definedbox}[label=pmc example1]{Example 1 of \benchmark}
    \textbf{Statement:} Josh decides to try flipping a house. He buys a house for \$80,000 and then puts in \$50,000 in repairs. This increased the value of the house by 150\%. How much profit did he make?
    \textcolor{teal}{\# Excepted Answer: 70,000}
    \tcbline
    \textbf{M Version:} Josh decides to try flipping a house. He buys a house for \$80,000 and then puts \textcolor{red}{\sout{\$50,000}} \textcolor{blue}{some cost} in repairs. This increased the value of the house by 150\%. How much profit did he make?
    \tcbline
    \textbf{C Version:} Josh decides to try flipping a house. He buys a house for \$80,000 and then puts in\$50,000 in repairs. This increased the value of the house by 150\%\textcolor{blue}{, but the market value of the house after repairs is only \$100,000}. How much profit did he make?
    \textcolor{teal}{(\# market value Contrary to the expected )}

    \end{definedbox}
\end{figure}

\begin{figure}[!hthb]
    \label{fig:pmc example2}
    \begin{definedbox}[label=pmc example2]{Example 2 of \benchmark}
    \textbf{Statement:} Janet’s ducks lay 16 eggs per day. She eats three for breakfast every morning and bakes muffins for her friends every day with four. She sells the remainder at the farmers' market daily for \$2 per fresh duck egg. How much in dollars does she make every day at the farmers' market?
    \textcolor{teal}{\# Excepted Answer: 14}
    \tcbline
    \textbf{M Version:} 
    Janet’s ducks lay 16 eggs per day. She eats \textcolor{red}{\sout{three}} \textcolor{blue}{some} for breakfast every morning and bakes muffins for her friends every day with four. She sells the remainder at the farmers' market daily for \$2 per fresh duck egg. How much in dollars does she make every day at the farmers' market?
    \tcbline
    \textbf{C Version:} Janet’s ducks lay 16 eggs per day. She eats three for breakfast every morning and bakes muffins for her friends every day with four. She sells the remainder at the farmers' market daily for \$2 per fresh duck egg. How much in dollars does she make every day at the farmers' market\textcolor{blue}{if she give 10 eggs away to her neighbor}?
    \textcolor{teal}{(\# She only left 9 eggs, can not give away 10 eggs)}

    \end{definedbox}
\end{figure}

\subsubsection{Constrction prompt}
The construction prompt we used is shown in the example 3,4,5.
\begin{figure}[!hthb]
    \label{fig:pmc construction prompt1}
    \begin{definedbox}[label=pmc construction prompt1 ]{Constrction prompt for missing type}
    Given the following math problem, identify all the variables and constraints involved. Then, modify the problem by replacing a key numerical value in one of the constraints with an indefinite placeholder (e.g., “some number”, “a certain value”, etc.), such that the resulting problem lacks sufficient information to determine a unique solution.

    You can answer with following step:\\
    Step 1: Variable and Constraint Identification.\\
    Step 2: Decide the mutated Variable or
    constraint and explain the reason.\\
    Step 3: Answer with final mutated problem.\\
    Original Problem:
    \{Problem\}
    
    Modified Problem:
    [Your answer]

    \end{definedbox}
\end{figure}

\begin{figure}[!hthb]
    \label{fig:pmc construction prompt1}
    \begin{definedbox}[label=pmc construction prompt1 ]{Constrction prompt for contra type}
     Given the following math problem, identify all the variables and constraints involved. Then, modify the problem by introducing an additional constraint that directly conflicts with an existing one. The resulting problem should contain contradictory information that makes it logically unsolvable.

    You can answer with following step:\\
    Step 1: Variable and Constraint Identification.\\
    Step 2: Decide the mutated Variable or
    constraint and explain the reason.\\
    Step 3: Answer with final mutated problem.\\
    Original Problem:
    \{Problem\}
    
    Modified Problem:
    [Your answer]

    \end{definedbox}
\end{figure}

\begin{figure}[!hthb]
    \label{fig:pmc example2}
    \begin{definedbox}[label=pmc construction prompt1 ]{Validation prompt}
Given the following math problem, determine whether it is solvable. If not, identify why the problem is ill-defined. Specifically, analyze whether the conditions provided are insufficient or self-contradictory, making it impossible to derive a unique solution.

You can answer with the following steps:  \\
Step 1: Variable and Constraint Identification.  \\
Step 2: Analyze whether the problem is solvable under the given constraints. If it is unsolvable, explain whether it is due to missing information or contradictory conditions, and identify the responsible part(s).  \\ 
Step 3: Give the final feedback if the question is unsolvable

Problem:  
\{Problem\}

Answer:  
[Your answer]

    \end{definedbox}
\end{figure}

\subsubsection{Human annotators}
When the LLM used for verification outputs inconsistent responses, we will enable human annotators to verify. Our annotators come from within the lab, no more than 5 master's and doctoral students.
\newpage

\subsection{Details of algorithm \algo}
\label{app:algo}
In this part, we will introduce the details of our algorithm \algo.
\subsubsection{Prompts in \algo}
We show the prompts we use in \algo with examples 6 and 7.

\begin{figure}
    \begin{definedbox}[label= fig: prompts used in algo]{prompts used in \algo -1}
    \textbf{Refine module prompt}\\
    \begin{small}
    I have previously asked you to write Z3 constraints for a problem. However, the current set of constraints for the variable may have omissions or errors. 
    I would like you to review it from the following two aspects and make appropriate modifications if necessary:\\
    1. Based on the problem description, consider whether the current constraints accurately capture the problem.\\
    2. Add constraints based on real-world knowledge, considering whether there are any missing modeling statements, such as the quantity of items should be >= 0, or the relationships between the sides of a triangle. \\
    Please note that you only need to add constraints to the CURRENT HEAD VARIABLE; in other words, the new constraints MUST include the head variable! \\
    You can first provide your thought process, and then write the new constraints that include the head variable after the identifier <SOS>\\

    You can follow the example:\\
    Question: Josh decides to try flipping a house. He buys a house for \$80,000 and then puts in \$50,000 in repairs. This increased the value of the house by 150\%, but the market value of the house after repairs is only \$100,000. How much profit did he make?\\
    Existing Constraints: \\
    solver.add(initial-cost == 80000)  \\
    solver.add(total-investment == initial-cost + repair-cost)   \\
    solver.add(repair-cost == 50000) \\
    solver.add(increased-value-percentage == 0.5)  \# 150\% increase \\
    solver.add(expected-value == initial-cost * (100 + increased-value-percentage)) \\   
    solver.add(market-value-after-repairs == 100000)\\
    solver.add(total-investment >= 0)\\
    solver.add(profit == market-value-after-repairs - total-investment)\\
    solver.add(profit >= 0)       \\
    solver.add(expected-value >= 0)\\
    Now head variable: expected-value\\
    Now existing constraints with head variable:\\
    solver.add(expected-value == initial-cost * (100 + increased-value-percentage))  \\
    solver.add(expected-value >= 0)\\
    Answer:\\  
    1. For the constraint expected-value == initial-cost * (100 + increased-value-percentage), the equation for expected-value in the problem should be initial-cost * (1 + increased-value-percentage). Therefore, this constraint should be modified to solver.add(expected-value == initial-cost * (1 + increased-value-percentage)).\\
    2. For the constraint solver.add(expected-value >= 0) aligns with real-world requirements. Additionally, since expected-value is an unknown variable, it is appropriate to add real-world constraints, so this should be retained.\\
    3. Furthermore, expected-value and market-value-after-repairs refer to the same entity in the problem, so a constraint should be added: market-value-after-repairs == expected-value.\\
    <SOS>\\
    So, new Constraints with head variable is \\
    solver.add(expected-value == initial-cost * (1 + increased-value-percentage)) \\
    solver.add(expected-value >= 0) \\
    solver.add(expected-value == market-value-after-repairs) \\  
    Question:\textcolor{red}{\{question\}}\\
    Existing Constraints:\textcolor{red}{\{constraint\}}\\
    Now head variable:\textcolor{red}{\{head\}}\\
    Now existing constraints with head variable:\textcolor{red}{\{constrain-head\}}\\
    Answer:
    \end{small}
    \end{definedbox}
\end{figure}

\begin{figure}
    \begin{definedbox}[label= fig: prompts used in algo2]{prompts used in \algo -2}
    \textbf{Verification module prompt}\\
    Please judge which set of constraints is better for the given problem, including all constraints of variable "X".\\
    Problem: \textcolor{red}{\{question\}} \\
    variable:\textcolor{red}{\{head\}} \\
    Constrains set1:\textcolor{red}{\{cons1\}} \\
    Constrains set1 ans:\textcolor{red}{\{cans1\}} \\
    Constrains set2:\textcolor{red}{\{cons2\}} \\
    Constrains set2 ans:\textcolor{red}{\{cans1\}} \\
    Please write down your thinking process first, and finally output, "I think Constrains set1 is better", or "I think Constrains set2 is better". 
    \end{definedbox}
\end{figure}

\subsubsection{Formal tools}
The SMT-LIB(Satisfiability Modulo Theories Library) ~\cite{barrett2010smt} is a tool for working with satisfiability problems. It provides a standard notation compatible input language for representing logical formulas. And powful SMT solvers, such as Z3~\cite{Moura2008z3}, extend the classical boolean satisfiability problem (SAT problem) to enable verification of numerical arithmetic problems, among others. The SMT solver will initially determine whether the modeled problem is satisfiable (SAT/UNSAT). If it is satisfiable, the solver will then provide a feasible solution within the feasible domain of the problem. Specifically, we use z3 as a formal tool in the paper.

\subsubsection{Double-check solving strategy with SMT solver}
We use a double-check strategy when checking with the SMT solver. Specifically, we verify both the satisfiability of the formal expression and the uniqueness of the solution. To be specific, to check the satisfiability of the formal expression, we utilize the Z3 solver. This strategy regards the problem as ill-defined and rejects the answer if the formal expression is unsatisfiable(UNSAT). 
To assess the uniqueness of the solution, We develop this check through a two-stage process.
First, we utilize the Z3 solver to determine one solution and subsequently incorporate this candidate solution as a constraint into the formal expression. If the formal expression remains satisfiable, then it implies that the formal expression encompasses multiple solutions, leading the strategy to reject the answer as it violates the uniqueness of the answer. 

To be precise, in the solution phase, our strategy let the SMT solver return four possible different values:
\begin{itemize}
    \item \textbf{Error}: Indicates that the modeling cannot be successfully completed. Similar to a compilation error, we do not consider it as a valid state.
\item \textbf{UNSAT}: Indicates that the modeling state cannot be satisfied, there are contradictory conditions, and the answer is rejected.
\item \textbf{Multi}: We believe that the question is ambiguous, resulting in multiple solutions, and the answer is rejected.
\item \textbf{Ans}: Returns a normal real number, representing the answer to the question.
\end{itemize}

\subsubsection{An example for \algo}

Our approach to determining variable-constraint relationships is as follows:

\begin{itemize}
  \item \textbf{Preparation Phase (Variables → Constraints)}: For a given variable, directly retrieve all constraints containing that variable from the constraint pool.
  \item \textbf{Update Phase (Constraints → Variables)}: For a given constraint, we identify all new associated variables in it.
\end{itemize}

To further illustrate this method, we present a concrete example using a contra-type problem in PMC (example 8) to demonstrate the search process:

\begin{figure}
    \begin{definedbox}[label= fig: examples used in algo2]{Example in \algo}

\textit{"Josh decides to try flipping a house. He buys a house for 80,000 \textbf{and then puts in} 50,000 in repairs. This increased the value of the house by 150\%, but the market value of the house is only \$100,000. How much profit did he make?"}

After the initialization step, we obtain an initial constraint system, represented in Python Z3 code. This system consists of a variable queue and a constraint pool.

\textbf{Variables:}

\begin{quote}
\texttt{"initial-cost", "repair-cost", "increased-value-percentage", "expected-value", "market-value-after-repairs", "profit", "total-investment"}
\end{quote}

\textbf{Constraints:}

\begin{verbatim}
initial-cost == 80000
repair-cost == 50000
market-value-after-repairs == 100000
increased-value-percentage == 0.5
total-investment == initial-cost + repair-cost
expected-value == initial-cost * (100 + increased-value-percentage)
profit == market-value-after-repairs - total-investment
\end{verbatim}

After the Initialization, assume that the first element in the variable queue is \texttt{"expected-value"}, we will demonstrate a single iteration of the search process.

\textbf{Preparation}

Identify constraints involving this variable \texttt{"expected-value"}:

\begin{verbatim}
expected-value == initial-cost * (100 + increased-value-percentage)
\end{verbatim}

\textbf{Exploration}

Utilize LLM knowledge to refine the constraints by generating a constraints set with the head variable \texttt{"expected-value"}:

\begin{verbatim}
expected-value == initial-cost * (basic_multiplier + increased-value-percentage)
basic_multiplier == 1
\end{verbatim}

\textbf{Verification}

Compare the original constraint system with the refined one and select the better version.

(In this case, the newly generated constraint set is selected).

\textbf{Update}

Replace the outdated constraint with the refined one.

Identify any newly introduced variables (e.g., \texttt{"basic\_multiplier"}) and append them to the tail of the variable queue for subsequent iterations.

\end{definedbox}
\end{figure}



\newpage
\subsection{Details of Experiment}
\label{app:expe}
In this section, we provide additional experimental details and discussions. However, due to time and computational constraints, some of the analyses were conducted on a subset of the dataset.
\subsubsection{Setup}
\textbf{Compared methods.} 
We selected three representative few-shot prompting methods, along with the zero-shot method that utilizes the intrinsic capabilities of the model, and compared them with our proposed \algo method. The methods are introduced as follows: 
(1)\textbf{Basic}, which is the zero-shot baseline method, directly feeds the problem and instructions to the LLMs without any example problem in the context. 
(2)\textbf{CoT}, ~\cite{wei2022cot}, requires the model to explicitly output intermediate step-by-step reasoning through natural language before providing the final answer.
(3)\textbf{PAL}~\cite{gao2023pal}, converts each step of problem-solving into a programming language format and subsequently utilizes an external programming language interpreter for execution, thereby obtaining the results.
(4)\textbf{Satlm}~\cite{ye2024satlm}, utilizes SMT-LIB to model the problems, then uses an external SMT solver to check for a feasible solution to the problem as well as obtain the ground-truth answer. 

\textbf{Prompts.} For the few-shot prompting methods, we prepared four contextual examples (4-shot) for each method, consisting of two well-defined problems and two ill-defined problems. In the system prompt, we explicitly informed the model about the potential presence of ill-defined problems. If the model determines that a problem is unsolvable, it is instructed to output a statement containing the term "unsolvable." This allows us to evaluate whether the model successfully identifies ill-defined problems.

\textbf{Set up details for Sec4.3.} At this part, we employed a balanced sampling strategy to fairly assess the ability to identify ill-posed problems and solve well-defined problems simultaneously. (with a solvable/unsolvable problem ratio of $\alpha= 1:1$), selecting 500 samples from the ill-defined problem set (Table \ref{Table:aood results}) and the well-defined problem set (Table \ref{Table:aid results}) to construct a 1000-sample test set. After three repeated experiments, we report the mean ± standard deviation in Table \ref{Table: real results}.

\subsubsection{Prompts used in Preliminary experiments}\
We show the prompts we use in preliminary experiments to reflect the trade-off dilemma with examples 9.
\begin{figure}[!htpb]
    \begin{definedbox}[label= fig: trade-off prompts]{prompts used in Preliminary experiments}
    \textbf{Pure prompt for ill-defined problem}\\Now we have some math problems that may be ill-defined. Please judge whether they are indeed ill-defined (no unique real number solution can be determined). 
    If there is indeed no solution, answer true, otherwise answer false. Explain the reason first and then answer.
    \tcbline
    \textbf{Pure prompt for well-behaved problem}\\
    You're an experienced elementary school teacher, and I'm now expecting you to solve some math problems.
    \tcbline
    \textbf{Mixed prompts}\\
    You're an experienced elementary school teacher, and I'm now expecting you to solve some math problems. \
        If you find these problems unsolvable, please output “this is unsolvable”. Or please solve this answer, and give the final answer with format "The answer is X"
    \end{definedbox}
\end{figure}

\subsubsection{More experiment results on other benchmark}
\begin{table}[!htpb] 
\centering 
\caption{R-Rate on MathTrap} 
\begin{tabular}{ccccc} 
\hline Model & Deepcoder & Qwen7b & Qwen3b & Qwen1.5b \\ 
\hline Zero & 22.95 & 15.57 & 15.57 & 13.72 \\ \hline Ours & 65.57 & 86.06 & 88.89 & 74.59 \\ 
\hline 
\end{tabular} 
\end{table}

Here, we also tested our method on several other benchmarks that involve refusal to answer. Our method also demonstrated superior performance on MathTrap. However, MathTrap's mathematical problems involve a significant amount of geometry and algebra, which are not well-suited for formal tool modeling. This is also not suitable for methods such as PAL. So we only compare ours with the zero-shot method. In such scenarios, our method adopts a relatively conservative approach, rejecting any problem it cannot confidently solve in order to maintain the safety of the reasoning system.

\subsubsection{More ablation about \algo}
We further conduct ablation of the algorithm from the following aspects:

\textbf{Variable Ordering.} By default, we refine variables in the order they appear in the problem statement. This approach often approximates the topological order of problem-solving steps, especially when the number of solution steps is limited. We also experimented with an alternative order: refining variables based on their frequency of occurrence in the constraints (from highest to lowest).

As shown in the table below, using the frequency-based iteration order improved performance on "contra" type problems while degrading it on "missing" type problems, with "well-defined" problems remaining stable. Our analysis suggests this is because "contra" problems often have contradictions embedded in hidden variables at constraint intersections (with high occurrences), which are more readily identified and optimized with frequency-based ordering. Conversely, in "missing" type problems, the missing variables frequently appear earlier in the data, making sequential iteration more effective.

\textbf{Iteration Strategy.} We experimented with the number of variable iterations as a hyperparameter T. While additional iterations bring slight performance improvements, they also introduce significant increases in computational and time costs. After weighing performance gains against resource consumption, we adopted a single-iteration setting as the most cost-effective choice. Detailed results are presented in the table below. (Due to time and computational limitations, we report results using the Qwen-7B model, based on 300 randomly sampled instances per dataset.)

\begin{table}[!htpb]
\centering
\caption{Performance under different configurations}
\label{table:config}
\begin{tabular}{lcccc}
\toprule
 & Default (T=1, appear-order) & T=2  & T=3  & frequency-order \\
\midrule
Contra   & 56.89 & 57.67 & 56.33 & 61.00 \\
Missing  & 92.60 & 93.00 & 91.67 & 87.67 \\
Well-type & 79.09 & 77.00 & 79.67 & 79.00 \\
\bottomrule
\end{tabular}
\end{table}

\subsubsection{Discussion about reasoning in realistic scenarios}
\textbf{Discussion of dataset ratios}

In our paper, we adopted a balanced setting($i.e., D_w : D_i = 1 : 1$) to measure the reaction score. This balanced approach allows us to evaluate the capability of methods to both answer well-defined problems and reject ill-defined problems with equal importance. This evaluation strategy is analogous to how imbalanced classification studies often report balanced metrics to properly assess model performance across all classes~\cite{thabtah2020imbalancesurvey}. By maintaining this balanced setting, we provide a more comprehensive and fair assessment of each method's capabilities of answering and rejecting. Additionally, we compared the R-score performance across different dataset ratios (defined as $\alpha = D_w : D_i$) on the Qwen1.5B model, and our method consistently demonstrated superior results.

\begin{table}[h!]
\centering
\caption{Performance among different data ratios}
\begin{tabular}{@{}lccccc@{}}
\toprule
$\alpha$ & 0.2 & 0.5 & 1 & 2 & 5 \\
\midrule
CoT & $44.61 \pm 1.02$ & $49.58 \pm 2.00$ & $53.91 \pm 1.16$ & $58.96 \pm 0.78$ & $62.83 \pm 1.55$ \\
CoT + Ours & $64.40 \pm 0.43$ & $64.05 \pm 0.60$ & $63.26 \pm 0.84$ & $64.33 \pm 0.89$ & $62.91 \pm 0.79$ \\
PAL & $16.01 \pm 0.66$ & $24.03 \pm 1.12$ & $32.85 \pm 1.00$ & $41.15 \pm 0.49$ & $49.53 \pm 2.89$ \\
PAL + Ours & $65.26 \pm 1.54$ & $62.46 \pm 0.22$ & $62.28 \pm 0.65$ & $58.55 \pm 1.13$ & $58.84 \pm 0.56$ \\
\bottomrule
\end{tabular}
\end{table}

\textbf{More convincing metrics}

To prevent excessive score inflation through question rejection (where rejecting all questions would yield only 50\% of the total score), we introduce the R*-score metric as below
\[
\frac{\sum_{p \in D_i} \mathbb{I}[f(p) = \text{Reject}] + \sum_{p \in D_w} \mathbb{I}[f(p) = g(p)]}{|D_i| + |D_w|}
\]

We evaluate our method under balanced settings and present the results in the following table. Our approach maintains superior performance in most scenarios(R*-score), demonstrating that our performance gains do not stem from simply rejecting most questions.

\begin{table}[h!]
\centering
\caption{Perfomance among R-score and R*-score}
\begin{tabular}{@{}lcccc@{}}
\toprule
 & \multicolumn{2}{c}{Qwen 1.5B} & \multicolumn{2}{c}{Qwen 3B} \\
\cmidrule(lr){2-3} \cmidrule(lr){4-5}
Method & R-score & R*-score & R-score & R*-score \\
\midrule
CoT & $53.91 \pm 1.16$ & $51.10 \pm 2.08$ & $65.93 \pm 0.73$ & $65.10 \pm 1.04$ \\
CoT + Ours & $63.26 \pm 0.84$ & $53.10 \pm 0.06$ & $73.98 \pm 0.28$ & $66.93 \pm 0.28$ \\
PAL & $32.85 \pm 1.00$ & $30.63 \pm 0.18$ & $48.56 \pm 0.22$ & $47.66 \pm 0.49$ \\
PAL + Ours & $62.28 \pm 0.65$ & $51.90 \pm 1.15$ & $74.08 \pm 1.17$ & $65.73 \pm 1.30$ \\
\bottomrule
\end{tabular}
\end{table}

\subsubsection{Performance on Larger LLM}
We've extended our experimental results to include GPT-4-0613 and GLM-4-Plus. These findings indeed demonstrate that larger language models exhibit a stronger ability to identify ill-defined problems. However, it's crucial to highlight that even with these powerful models, our proposed method can further enhance their capability to recognize and handle such issues, significantly improving their robustness, particularly in "contra" type problems.

\begin{table}[h]
\centering
\begin{tabular}{l l c c}
\hline
Model & Dataset & Basic & Ours \\
\hline
gpt4-0613 & Contra  & 35.00 & 71.00 \\
          & Missing & 82.00 & 85.00 \\
GLM-4-Plus & Contra  & 44.67 & 64.00 \\
           & Missing & 73.67 & 86.00 \\
\hline
\end{tabular}
\caption{Performance comparison between baseline and our method.}
\end{table}

Even though the ability of large LLMs to handle robust mathematical reasoning has improved, this doesn't diminish the importance of addressing ill-defined problems. In reality, not all application scenarios possess the resources or infrastructure to deploy ultra-large models. For instance, mid-sized models (around 7 billion parameters) are widely adopted in practical applications due to their excellent deployability and cost-effectiveness. In these contexts, robust reasoning capabilities remain a critical focus.

Furthermore, we contend that symbolic methods offer a unique advantage in this scenario, rather than being overly complex. Specifically, a symbolic solver can be effectively utilized as a tool to assess the completeness of problem descriptions, while the language model focuses on modeling the constraint system. This division of labor frees the model from the trade-off between identifying pathological problems and solving well-defined ones. And even with advanced larger LLMs, experiments show they can't fully escape this trade-off, yet our symbolic approach still proves effective.

\subsubsection{Computational cost discussion}
Our framework employs a sequential iterative structure. In terms of memory and computational costs, it's comparable to standard LLM inference, as most resource consumption doesn't significantly increase. The primary overhead lies in runtime. We use the Z3 SMT solver, which has a relatively low CPU footprint. For example, when running Qwen-7B on our proposed PMC dataset with an RTX 4090, GPU and CPU memory consumption are approximately 15 GB and 1.6 GB, respectively.

While our method does have a higher runtime than zero-shot/few-shot inference, this is a common characteristic of test-time scaling approaches. Our proposed method achieves linear time complexity with respect to the number of variables, rather than exponential growth. Our empirical runtime measurements confirm this linear scaling behavior, significantly outperforming methods that build constraint systems using tree search. The table below shows the average time required for the Qwen-7B model to solve a single problem using different methods.

\begin{table}[h]
\centering
\begin{tabular}{l c c c c}
\hline
Method & Basic & SMT & Ours & Tree Search \\
\hline
Time   & 6.6s  & 12s & 79.8s & 156s \\
\hline
\end{tabular}
\caption{Time consumption comparison of different methods.}
\end{table}

\subsubsection{LLM verification in VCSearch}
We acknowledge the current limitations of Large Language Models (LLMs) in modeling symbolic systems. However, it's important to clarify that our LLM-Judge relies not solely on the model's output. To enhance the reliability of its judgments, we also provide the problem's text description and the SMT solver's execution results (Equation 7) as additional inputs.

Given that incorporating human evaluation into every LLM-Judge process would demand substantial annotation resources and time, it's impractical for real-world application. Therefore, we propose an alternative metric: \textbf{Judge-Error-Rate (JER)}. Among samples where the final output was incorrect, any correct answer that was found during the search process but not ultimately selected as the final output is counted as a judge error. We calculate JER as the proportion of these judge errors among all judging instances. This metric serves as an effective measure of LLM-Judge's reliability. We've calculated the JER for our method across various models and datasets, and the results demonstrate the high effectiveness of our LLM-Judge approach.

\begin{table}[!htpb]
\centering
\caption{JER on different datasets}
\label{table:contra-missing}
\begin{tabular}{lccc}
\toprule
Model   & Contra & Missing & Well-defined \\
\midrule
Qwen7b   & 8.80  & 1.11 & 1.69 \\
Qwen3b   & 11.64 & 2.60 & 1.68 \\
Qwen1.5b & 1.40  & 9.26 & 2.44 \\
\bottomrule
\end{tabular}
\end{table}

\end{document}